\documentclass[conference]{IEEEtran}
\IEEEoverridecommandlockouts
\usepackage{cite}
\usepackage{amsmath,amssymb,amsfonts}
\usepackage{algorithmic}
\usepackage{graphicx}
\usepackage{textcomp}
\usepackage{xcolor}
\usepackage{subfigure}
\usepackage{multirow}
\usepackage{booktabs}
\usepackage{amsmath,bm}
\usepackage[inline]{enumitem}
\usepackage{color}
\usepackage{threeparttable}
\usepackage{url}

\makeatletter
\newcommand{\linebreakand}{%
\end{@IEEEauthorhalign}
\hfill\mbox{}\par
\mbox{}\hfill\begin{@IEEEauthorhalign}
}
\makeatother

\def\BibTeX{{\rm B\kern-.05em{\sc i\kern-.025em b}\kern-.08em
    T\kern-.1667em\lower.7ex\hbox{E}\kern-.125emX}}
    
\newcommand{\para}[1]{\noindent\textbf{#1}}
\newcommand{\real}[1]{\mathbb{R}^{#1}}

\usepackage{array}
\newcolumntype{H}{>{\setbox0=\hbox\bgroup}c<{\egroup}@{}}

\newcounter{sancounter}
\DeclareRobustCommand{\san}[1]{\textbf{\color{blue}/* #1 (san) */}\stepcounter{sancounter}\typeout{LaTeX Warning: sandipan comment \thesancounter: #1 (line \the\inputlineno)}}
\newcounter{hucounter}
\DeclareRobustCommand{\hussain}[1]{\textbf{\color{red}/* #1 (hussain) */}\stepcounter{hucounter}\typeout{LaTeX Warning: hussain comment \thehucounter: #1 (line \the\inputlineno)}}
\newcounter{rkcounter}
\DeclareRobustCommand{\rkern}[1]{\textbf{\color{orange}/* #1 (roman) */}\stepcounter{rkcounter}\typeout{LaTeX Warning: roman comment \therkcounter: #1 (line \the\inputlineno)}}
\newcounter{mengcounter}
\DeclareRobustCommand{\meng}[1]{\textbf{\color{magenta}/* #1 (meng) */}\stepcounter{mengcounter}\typeout{LaTeX Warning: meng comment \thesancounter: #1 (line \the\inputlineno)}}
\newcounter{elcounter}
\DeclareRobustCommand{\el}[1]{\textbf{\color{cyan}/* #1 (elex) */}\stepcounter{elcounter}\typeout{LaTeX Warning: el comment \thesancounter: #1 (line \the\inputlineno)}}
\newcounter{dhecounter}
\DeclareRobustCommand{\dhe}[1]{\textbf{\color{green}/* #1 (dhe) */}\stepcounter{dhecounter}\typeout{LaTeX Warning: dhe comment \thedhecounter: #1 (line \the\inputlineno)}}
\DeclareRobustCommand{\mst}[1]{\textbf{\color{teal}/* #1 (mst) */}\stepcounter{mstcounter}\typeout{LaTeX Warning: mst comment \thedhecounter: #1 (line \the\inputlineno)}}    
    
\begin{document}

\title{Adversarial Inter-Group Link Injection Degrades the Fairness of Graph Neural Networks
}

\author{\IEEEauthorblockN{Hussain Hussain\textsuperscript{*}}
\IEEEauthorblockA{\textit{Graz University of Technology}}
\IEEEauthorblockA{
\textit{Know-Center GmbH}\\
hussain@tugraz.at}
\and
\IEEEauthorblockN{Meng Cao\textsuperscript{*}}
\IEEEauthorblockA{\textit{Nanjing University} \\
caomeng@smail.nju.edu.cn}
\and
\IEEEauthorblockN{Sandipan Sikdar}
\IEEEauthorblockA{\textit{L3S Research Center} \\
sandipan.sikdar@l3s.de}
\and
\IEEEauthorblockN{Denis Helic}
\IEEEauthorblockA{\textit{Modul University Vienna}}
\IEEEauthorblockA{\textit{Graz University of Technology}\\
denis.helic@modul.ac.at}
\and
\IEEEauthorblockN{Elisabeth Lex}
\IEEEauthorblockA{\textit{Graz University of Technology} \\
elisabeth.lex@tugraz.at}
\and
\IEEEauthorblockN{Markus Strohmaier}
\IEEEauthorblockA{\textit{University of Mannheim}}
\IEEEauthorblockA{\textit{GESIS - Leibniz Institute for the Social Sciences}}
\IEEEauthorblockA{\textit{CSH Vienna} \\
markus.strohmaier@uni-mannheim.de}
\and
\IEEEauthorblockN{Roman Kern}
\IEEEauthorblockA{\textit{Graz University of Technology}}
\IEEEauthorblockA{
\textit{Know-Center GmbH}\\
rkern@tugraz.at}
}
\maketitle

\begingroup\renewcommand\thefootnote{*}
\footnotetext{Equal contribution.}
\endgroup
\begingroup\renewcommand\thefootnote{}
\footnotetext{Accepted as a short paper at ICDM 2022.\\
\copyright IEEE 2022. Personal use of this material is permitted. Permission from IEEE must be obtained for all other uses, in any current or future media, including reprinting/republishing this material for advertising or promotional purposes, creating new collective works, for resale or redistribution to servers or lists, or reuse of any copyrighted component of this work in other works.}

\endgroup

\begin{abstract}
We present evidence for the existence and effectiveness of adversarial attacks on graph neural networks (GNNs) that aim to degrade fairness. These attacks can disadvantage a particular subgroup of nodes in GNN-based node classification, where nodes of the underlying network have sensitive attributes, such as race or gender. We conduct qualitative and experimental analyses explaining how adversarial link injection impairs the fairness of GNN predictions. For example, an attacker can compromise the fairness of GNN-based node classification by injecting adversarial links between nodes belonging to opposite subgroups and opposite class labels. Our experiments on empirical datasets demonstrate that adversarial fairness attacks can significantly degrade the fairness of GNN predictions \emph{(attacks are effective)} with a low perturbation rate \emph{(attacks are efficient)} and without a significant drop in accuracy \emph{(attacks are deceptive)}. This work demonstrates the vulnerability of GNN models to adversarial fairness attacks. We hope our findings raise awareness about this issue in our community and lay a foundation for the future development of GNN models that are more robust to such attacks.
\end{abstract}


\section{Introduction}



\begin{figure}[t]
    \centering
    \includegraphics[width=.99\linewidth]{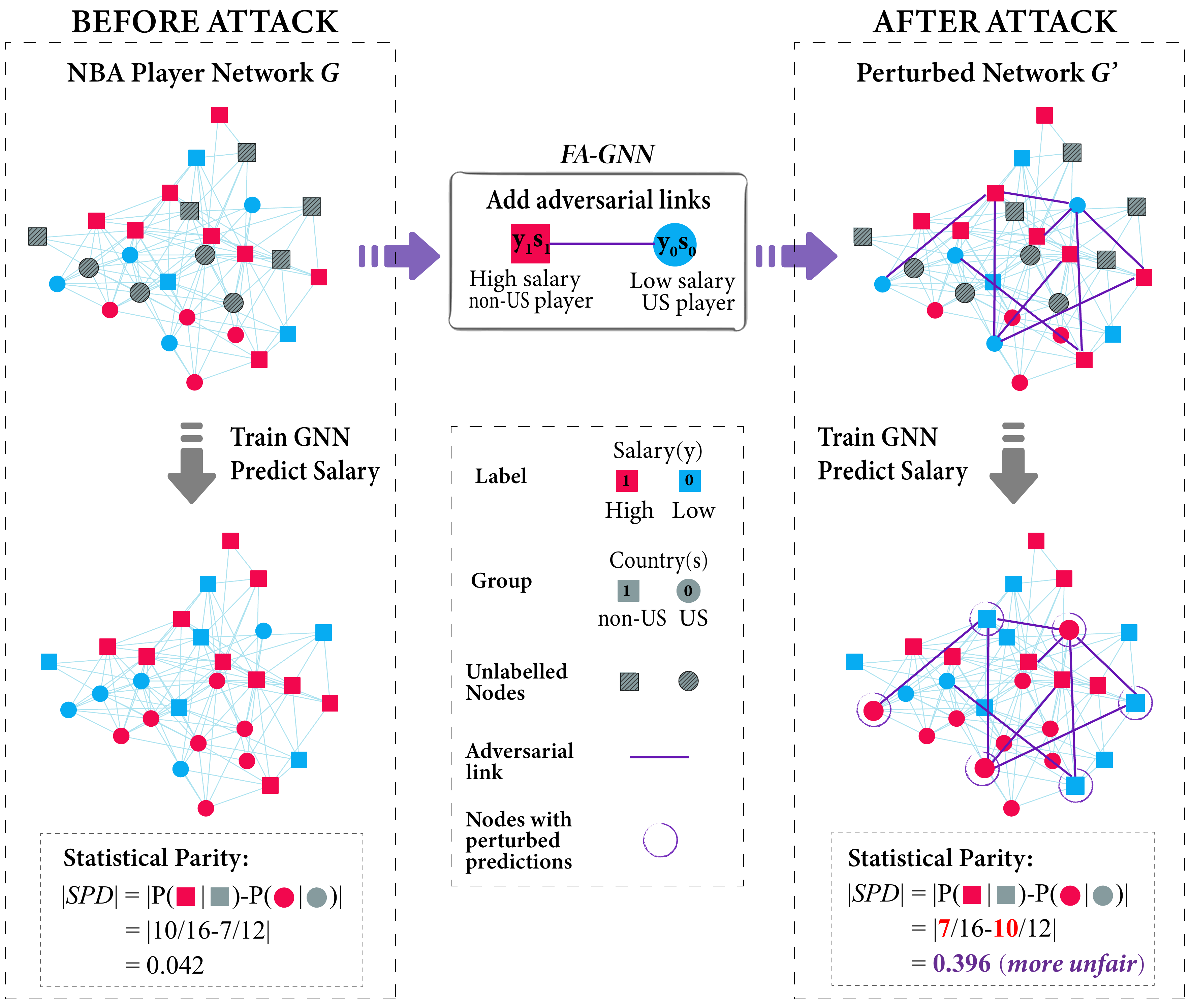}
    \caption{Fairness Attacks on GNNs (FA-GNN). In the NBA player network~\cite{dai2021say}, the player's nationality represents their sensitive attribute $s$: the US (round) or non-US (square). The classification task is to predict their salary level $y$: \textcolor[RGB]{242,7,78}{high} or \textcolor[RGB]{7,172,242}{low}.
    Suppose the attacker has access to the network $G$, with the sensitive attributes $s$ and the labels $y$ on a subset of nodes in $G$.
    The attacker can predict the labels for the rest of the nodes via a surrogate GNN~\cite{zugner2019adversarial}.
    Then one strategy of FA-GNN is to perturb $G$ by adding links between players with opposite labels and opposite sensitive attributes, e.g., high salary non-US players ($y_1s_1$) and low salary US players ($y_0s_0$).
    After training a GNN model on the perturbed network $G'$, the predictions of the nodes in purple circles differ from the predictions on the clean graph $G$, and the statistical parity largely increases. The GNN model becomes more unfair in that it is more likely to predict high salaries for US players. 
    This example illustrates how adversarial fairness attacks can degrade the fairness of GNN's predictions via inter-group link injection.
    }
    \label{fig:key}
\end{figure}

Previous research has already shown that graph neural networks (GNNs) are susceptible to adversarial attacks targeting prediction accuracy~\cite{zugner2018adversarial,dai2018adversarial,jin2020adversarial}.
Besides, in classical machine learning (ML) settings, recent studies have also uncovered various possibilities for adversarial attacks to compromise the \emph{fairness} of ML models~\cite{mehrabi2022}.
However, we still lack a deeper insight into GNNs' susceptibility to adversarial attacks on prediction fairness as well.
Here, we think of prediction fairness as a similar or equal treatment of (social) subgroups for a given prediction task. For example, in the job-hunting market where candidates are socially connected, employers use graph machine learning models (GNNs) to decide the salary of each candidate. Attackers from a specific demographic group, e.g. a certain race, gender, or religion, can intentionally conduct adversarial attacks, with the aim of making the model more likely to predict high salary for their own group, while severely hurting the benefit of the other groups (cf. Fig.~\ref{fig:key}).
Hence, adversarial attacks on the fairness of GNNs enable an attacker to deliberately put individuals belonging to a subgroup of nodes at a disadvantage.
We thus turn our attention to the following problem:


\para{Problem.}
We define and analyze adversarial attack strategies that degrade the fairness of GNN-based node classifiers. 




\para{Method.}
We base our research on two separate yet related streams of research: adversarial attacks on GNNs that aim to reduce GNN classification accuracy~\cite{zugner2018adversarial,zugner2019adversarial, dai2018adversarial,ma2019attacking,wang2019attacking,wu2019adversarial}, and attacks on fairness in the context of classical machine learning ~\cite{mehrabi2020exacerbating,nanda2021fairness,chhabra2021fairness,solans2020poisoning}.
We measure fairness in terms of statistical parity, equality of opportunity and equalized odds, which capture the level of independence between predictions and group memberships.
In our node classification problem, we assume a binary class label $y$ with negative ($y$=0) or positive ($y$=1) values.
Each node also has a binary sensitive attribute $s$ representing its group: privileged ($s$=0) or unprivileged ($s$=1).
First, we formally define the problem of Fairness Attacks on GNNs (FA-GNN).
We then present four adversarial linking strategies based on injecting adversarial links between two subsets of nodes (from a specific class label and sensitive attribute value pair).
Through qualitative analysis, we develop insights on the influence of these linking strategies on the statistical parity of GNN predictions.
We further illustrate these strategies on synthetically generated graphs by evaluating the prediction fairness of the graph convolution network (GCN) model~\cite{kipf2016semi}.
Finally, we evaluate FA-GNN adversarial linking strategies on three real-world social network datasets.
Fig.~\ref{fig:key}, illustrates an FA-GNN strategy on the NBA player social network~\cite{dai2021say}, where the attack degrades the fairness of GNN predictions.

\para{Results.}
Our qualitative analysis and experimental evaluation demonstrate the existence and success of adversarial attacks on the fairness of GNN-based classifiers. 
In particular, our results suggest that adding edges between nodes belonging to opposite groups and opposite class labels leads to less fair predictions.
This finding is particularly interesting since it appears plausible that intra-group links promote fairness~\cite{spinelli2021fairdrop,masrour2020bursting}.
As one would expect, compared to our presented fairness attacks, both basic and state-of-the-art attacks on GNNs -- originally developed to reduce prediction accuracy -- do not degrade GNN's fairness.

\para{Contributions and implications.} 
    First, we present the problem of adversarial attacks on GNNs that can impair the fairness of node classification.
    Second, we provide theoretical insights on the consequences of adversarial link injection on the fairness of GNN-based node classifiers.
    Third, 
    through experiments on three real-world datasets, we demonstrate the effectiveness, efficiency, and deceptiveness of fairness attacks\footnote{We provide our code and a detailed hyperparameter description at \url{https://github.com/mengcao327/attack-gnn-fairness}}.
%
Our work raises awareness about the vulnerability of GNNs to fairness attacks and argues for the development of models that are more robust against such attacks.



\section{Preliminaries}\label{preliminaries}

\para{Notation.}
We consider a graph $G(V,E,X,y,s)$, where $V = \{1,...,N\}$ is the set of nodes, $E \subseteq V \times V$ is the set of edges, $X \in \real{N \times F}$ is the feature matrix, while $y \in \{0,1\}^N$ represent the node labels and $s \in \{0,1\}^N$ represent the node sensitive attributes.
Hence, each node $u \in V$ has a label $y(u)$ denoting its \textit{class}, a sensitive attribute (e.g., race, gender, etc.) $s(u)$, whose value determines its group membership, and an $F$-dimensional feature vector $X(u)$.

\para{Subsets.}
We consider a partition $\mathcal{S}$ of the node set $V$ into four subsets:
$\mathcal{S} = \{y_0s_0,y_0s_1,y_1s_0,y_1s_1\}$, where $y_is_j$ denotes the subset of nodes with value pair $y(.)=i$ \textit{and} $s(.)=j$, with 
$n_{y_is_j}$ representing its cardinality.
We overload this notation to represent the
number of nodes in a group $s=j$ as $n_{s_j}$.
Hereafter, the term \textit{subset} denotes nodes with a particular combination of label and sensitive attribute values, unlike the term \textit{group} that denotes nodes with a particular sensitive attribute value.

\para{Graph neural networks for node classification.}
GNNs~\cite{wu2020comprehensive} are machine learning models that operate on graph data.
GNNs update a node's representation by aggregating the initial representations of its neighbors, which is known as message passing and aggregation.
In (semi-)supervised node classification, we train a GNN on the ground truth classes of labeled nodes $V_l \subseteq V$ via backward propagation.
For binary node classification, the GNN outputs the predicted probability of each node belonging to class $y=1$, i.e., $GNN : G(V,E,X) \rightarrow [0,1]^N$.
After thresholding, this results in one prediction per node $\hat{y}(u) \in \{0,1\}$.

\para{Structural adversarial attacks on GNNs.}
Given a \emph{clean graph} $G(V,E,X,y,s)$, an adversarial attack~\cite{jin2020adversarial} modifies this graph and returns a \emph{perturbed graph} $G'$.
Structural adversarial attacks particularly change the clean edge set into a perturbed edge set $E'$.
However, the number of modifications that an attacker can make on the edge set is bounded by a perturbation rate $\delta$, i.e., the attacker can add or remove up to $\delta |E|$ edges. 
We further consider that the attacker modifies the graph prior to the GNN training and evaluation (poisoning attack).
Such adversarial attacks increase the prediction error of the GNN on the perturbed graph, particularly the error on nodes directly targeted by the attack~\cite{zugner2018adversarial}. We name the influenced GNN model as the \emph{victim GNN model}.


\para{Fairness metrics.}
We measure the performance of FA-GNN linking strategies in terms of three group fairness metrics, which capture the difference in predictive outcomes and error rates between the two groups.
To recognize which group is at an advantage or a disadvantage, we use a signed difference for the following metrics instead of the absolute difference used in~\cite{beutel2017data}.
For all three metrics, a positive value indicates that the GNN predicts $\hat{y}=1$ for group $s_0$ more often than $s_1$ and vice versa.

\emph{Statistical parity difference}~\cite{dwork2012fairness} measures the difference in predictive outcomes between the two groups
\begin{equation}
\label{eq:signed-parity}
    SPD = P(\hat{y} = 1 | s = 0) - P(\hat{y} = 1 | s = 1).
\end{equation}

\emph{Equality of opportunity difference}~\cite{hardt2016equality} measures the difference of the classifier's true positive rates (or equivalently, false negative rates~\cite{verma2018fairness}) between the two groups
\begin{equation}
    EOD = P(\hat{y} = 1 | y=1, s = 0) - P(\hat{y} = 1 | y=1, s = 1).
\end{equation}
%

\emph{Equalized odds difference}~\cite{hardt2016equality} measures the overall differences of the classifier's true positive rates and false positive rates (or equivalently, true and false negative rates) between the two groups. 
\begin{equation}
\begin{split}
    EQD & = P(\hat{y} = 1 | y=1, s = 0) - P(\hat{y} = 1 | y=1, s = 1) \\
    & + P(\hat{y} = 1 | y=0, s = 0) - P(\hat{y} = 1 | y=0, s = 1).
\end{split}
\end{equation}

\section{Fairness attacks on GNNs}\label{sec:fagnn}
In this section, we devise and analyze fairness attacks on GNNs via adversarial link injection.
First, we formalize the attack settings and describe adversarial linking strategies that work under these settings.
Then, we provide a qualitative analysis to understand the consequences of this adversarial linking on the fairness of GNNs.
Finally, we illustrate these consequences on synthetically generated graphs.

\subsection{Problem formulation}
\label{sec:formulation}
In this work, we study fairness attacks on GNN-based binary node classification with the following settings.

\para{Attacker's goal.}
The attacker's goal is to degrade the prediction fairness of a \emph{victim GNN} binary node classifier by increasing the absolute statistical parity difference $|SPD|$.
The attacker can achieve that by increasing or decreasing the (signed) $SPD$ depending on its initial value in the clean graph.
We elaborate more on the initial $SPD$ value and the suitable strategies in Section~\ref{sec:attack-outcome}. 

\para{Attacker's knowledge.}
The data available to the attacker include the structural information, feature information and sensitive attributes of the entire graph as well as the ground-truth labels of some nodes $W \subseteq V$, that is $(V,E,X,y(W),s)$. 
For example, in the job-hunting market, an attacker may have access to the candidates' attributes and relationships in the social network, as well as the demographic group of each candidate, but can only obtain the salary of some candidates.
The attacker has no access to the parameters or hyperparameters of the victim GNN model.
However, similar to the settings in~\cite{zugner2018adversarial, zugner2019adversarial}, the attack uses a \emph{surrogate GCN}~\cite{kipf2016semi} model to predict labels for the unlabeled nodes, which makes the attack a grey-box attack~\cite{jin2020adversarial}.
We denote the predictions of the surrogate GCN as $\bar{y}$ to distinguish them from the predictions of the victim GNN $\hat{y}$.

\para{Attacker's capabilities.}
The attacker has the capability of injecting edges into the graph up to a budget, defined by a perturbation rate $\delta > 0$, i.e., the attacker can add up to $\delta |E|$ edges to the graph.




\subsection{Attack strategy}
\label{sec:strategy}
We now introduce adversarial linking strategies for fairness attacks on GNNs (FA-GNN) that work within our settings.
These linking strategies take a clean graph $(V,E,X,y({W}),s)$ as an input and adds edges between random nodes from subset $A\in\mathcal{S}$ and random nodes from subset $B\in\mathcal{S}$ up to a perturbation rate $\delta$.
To be able to partition the graph, the attacker first trains a surrogate GCN model to predict the labels for \textit{unlabeled nodes} $\bar{y}(V \setminus W)$.
The attacker then sets the predictions of \textit{labeled nodes} as the known ground-truth labels $\bar{y}(W):=y(W)$.
This provides an approximation of which nodes belong to $A$ and $B$.
Finally, the attacker adds $\lfloor\delta |E|\rfloor$ edges between random nodes from $A$ and $B$.

\begin{figure}[t]
    \begin{flushright}
    \includegraphics[width=.9\linewidth]{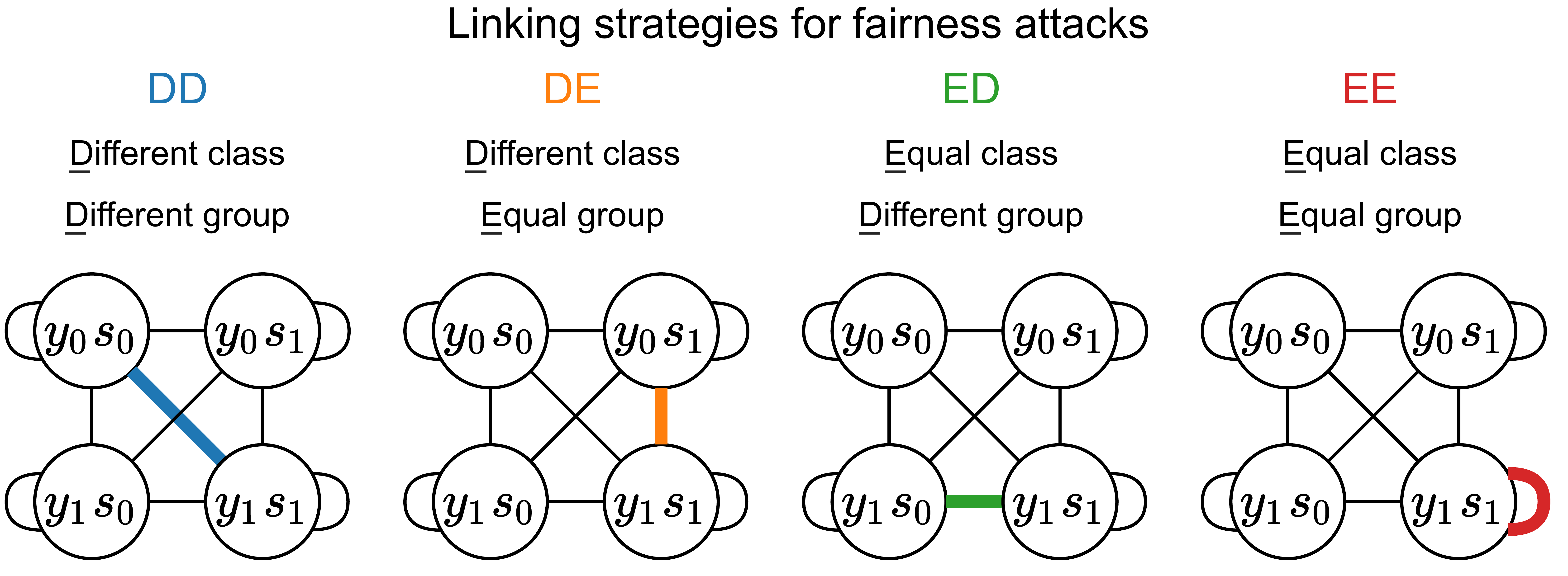}
    \end{flushright}
    \centering
    \includegraphics[width=\linewidth]{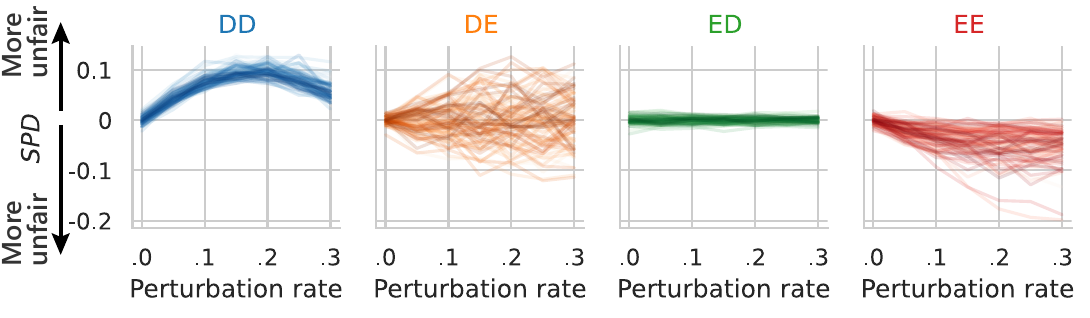}
    \caption{(Top) Linking strategies for fairness attacks involving nodes with $y=1,s=1$.
    Each circle $y_is_j$ represents the subset of all nodes with $y=i,s=j$.
    The colored line between two subsets denotes that the respective strategy adds links between random nodes of these subsets.
    (Bottom) Consequences of fairness attacks on synthetic graphs. 
    Each figure shows the statistical parity difference ($SPD$) of a linking strategy on synthetic graphs. The arrows show where the result becomes more unfair.
    We see that the DD strategy increases $SPD$ and the EE strategy reduces $SPD$.
    Meanwhile, ED fluctuates around 0, and DE also fluctuates around 0 with a high variance.
    The simulation results show that, DD and EE degrade the GNN fairness effectively, while DE is inconsistent and ED is ineffective.}
    \label{fig:linking}
\end{figure}

Choosing $A,B$ reduces to one of the following four linking strategies (illustrated in Fig.~\ref{fig:linking} - Top) that connect nodes of 
\begin{enumerate}
    \item \textcolor[RGB]{31,119,180}{DD}: \underline{D}ifferent class and \underline{D}ifferent group,
    \item \textcolor[RGB]{255,127,14}{DE}: \underline{D}ifferent class and \underline{E}qual group,
    \item \textcolor[RGB]{44,160,44}{ED}: \underline{E}qual class and \underline{D}ifferent group, or
    \item \textcolor[RGB]{214,39,40}{EE}: \underline{E}qual class and \underline{E}qual group, that is $A=B$.
\end{enumerate}
Note that in this paper we demonstrate only one possible way of performing a fairness attack, e.g., via \emph{adversarial link injection}. We leave other types of graph adversarial attacks on fairness for future work. 
Hereafter, we mainly use the term FA-GNN to refer to the aforementioned linking strategies.

\subsection{Attack consequences}
\label{sec:theoretical-analysis}


\begin{table}[b]
\caption{Binary classification confusion matrix with a binary sensitive attribute.}
\label{tab:confusion-matrix}
\small
\begin{tabular}{ll|ll|ll}
                            & Group                 & \multicolumn{2}{c|}{$s=0$}                             & \multicolumn{2}{c}{$s=1$}                             \\
                            & Class                 & \multicolumn{1}{c}{$y=0$} & \multicolumn{1}{c|}{$y=1$} & \multicolumn{1}{c}{$y=0$} & \multicolumn{1}{c}{$y=1$} \\ \hline
\multirow{2}{*}{Prediction} & $\hat{y}=0$           & $TN_{s_0}$                & $FN_{s_0}$                 & $TN_{s_1}$                & $FN_{s_1}$                \\
                            & $\hat{y}=1$           & $FP_{s_0}$                & $TP_{s_0}$                 & $FP_{s_1}$                & $TP_{s_1}$                \\ \hline
                      & \multicolumn{1}{c|}{$\sum$} & $n_{y_0s_0}$              & $n_{y_1s_0}$               & $n_{y_0s_1}$              & $n_{y_1s_1}$              \\ \hline
                  & \multicolumn{1}{c|}{$\sum$} & \multicolumn{2}{c|}{$n_{s_0}$}                         & \multicolumn{2}{c}{$n_{s_1}$}                        
\end{tabular}
\end{table}

We investigate the effects of the different adversarial linking strategies on GNN fairness in terms of statistical parity difference ($SPD$). As $SPD$ captures the predictive difference between the groups, we study the attack effects in terms of the classification errors for those groups (cf. binary classification confusion matrix in Table~\ref{tab:confusion-matrix}).
First, we rewrite Equation~\ref{eq:signed-parity} in terms of our four subsets:
\begin{equation}
\begin{split}
    SPD & = P(\hat{y} = 1, y = 0 | s = 0) + P(\hat{y} = 1, y = 1 | s = 0)\\
    & - P(\hat{y} = 1, y = 0 | s = 1) - P(\hat{y} = 1, y = 1 | s = 1).
\end{split}
\end{equation}
Next, we rewrite with the terms from the confusion matrix:
\begin{equation}
\begin{split}
    SPD &= \frac{FP_{s_0}+TP_{s_0}}{n_{s_0}} - \frac{FP_{s_1}+TP_{s_1}}{n_{s_1}},\\
\end{split} 
\end{equation}
where $FP_{s_j}$ and $TP_{s_j}$ are the false and true positive counts for group $s=j$.
Lastly, we rewrite once more using only error terms:
\begin{equation}
\label{eq:sp-confusion}
\begin{split}
    SPD &= \frac{FP_{s_0}-FN_{s_0}}{n_{s_0}} - \frac{FP_{s_1}-FN_{s_1}}{n_{s_1}} + \frac{n_{y_1s_0}}{n_{s_0}}-\frac{n_{y_1s_1}}{n_{s_1}},
\end{split}
\end{equation}
where $FN_{s_j}$ and $FP_{s_j}$ are the false negative and false positive counts in group $s=j$.
If we consider a random classifier or a constant classifier, the terms in Equation~\ref{eq:sp-confusion} cancel out and $SPD = 0$.
In case of a perfect classifier, all error terms vanish in Equation~\ref{eq:sp-confusion}, and we get $SPD = n_{y_1s_0}/n_{s_0}-n_{y_1s_1}/n_{s_1}$.
Hence, a perfect predictor is as fair as the original label distribution.

We use 
Equation~\ref{eq:sp-confusion} to investigate the different linking strategies of FA-GNN.
We explore homophilic graphs in our analysis (where likes attract), as GNNs are typically deployed on homophilic graphs~\cite{zhu2020beyond}.
According to previous work~\cite{zhang2020gnnguard}, adding edges between nodes of different classes increases the error rates on these nodes.
Conversely, adding edges between nodes of the same class decreases the error rates on these nodes.
We neglect the changes of errors on nodes that are not involved in the linking, since directly attacked nodes usually exhibit a higher error~\cite{zugner2018adversarial} (we address this assumption later in Section~\ref{sec:error}).
After performing the attack on the clean graph $G$, we get a new statistical parity difference on the perturbed graph $G'$, that is $SPD' = SPD + \Delta SPD$.
We denote the change in the number of false positives and false negatives on $s_j$ as $\Delta FP_{s_j}$ and $\Delta FN_{s_j}$, respectively.

\para{Analysis.}
Without loss of generality, we discuss the adversarial linking strategies from the perspective of nodes in $A = y_1s_1$.

    \underline{\textit{Case DD: $B = y_0s_0$.}} Here, we link nodes from different classes, so error rates would increase:
    $$\Delta FP_{s_0} \geq 0, \Delta FN_{s_1} \geq 0, \Delta SPD = \frac{\Delta FP_{s_0}}{n_{s_0}}+\frac{\Delta FN_{s_1}}{n_{s_1}}\geq 0$$
    Therefore, this linking strategy results in an \emph{attack against group $s_1$}.
    Note that we neglect $\Delta FP_{s_1}$ and $\Delta FN_{s_0}$ since they are not directly affected by the attack.
    Next, we discuss the remaining adversarial linking strategies of FA-GNN in the same manner.
    
    \underline{\textit{Case DE: $B = y_0s_1$.}} Here we also link nodes from different classes, so error rates would increase:
    $$\Delta FP_{s_1} \geq 0, \Delta FN_{s_1} \geq 0, \Delta SPD = \frac{\Delta FN_{s_1}}{n_{s_1}}-\frac{\Delta FP_{s_1}}{n_{s_1}}$$
    The sign of $\Delta SPD$ depends on the values of $\Delta FN_{s_1}$ and $\Delta FP_{s_1}$.
    This linking strategy results in a change on $SPD$ but the direction is not predictable, so the \emph{attack is not targeted}, that is, it does not specifically increase or decrease $SPD$.
    
    \underline{\textit{Case ED: $B = y_1s_0$.}} We link nodes from the same class, so error rates would decrease:
    $$\Delta FN_{s_0} \leq 0, \Delta FN_{s_1} \leq 0, \Delta SPD = \frac{\Delta FN_{s_1}}{n_{s_1}}-\frac{\Delta FN_{s_0}}{n_{s_0}}$$
    The sign of $\Delta SPD$ is unpredictable, hence the \emph{attack is not targeted}. 
    In the case of a small error rate on the clean graph, this linking strategy will not be effective, as $\Delta FN_{s_0}$ and $ \Delta FN_{s_1}$ will be small.
    
    \underline{\textit{Case EE: $B = y_1s_1$.}} Here we also link nodes from the same class, so error rates would decrease:
    $$\Delta FN_{s_1} \leq 0, \Delta SPD = \frac{\Delta FN_{s_1}}{n_{s_1}} \leq 0$$
    $SPD$ decreases, therefore this linking strategy results in an \emph{attack in favor of group $s_1$}.

In summary, linking $y_1s_1$ and $y_0s_0$ (nodes from different classes and different groups, that is DD) results in an attack targeted against $s_1$.
Besides, increasing the intra-connectivity of $y_1s_1$ (that is, EE) results in a targeted attack against $s_0$.
EE only manipulates one subset, which might not be as effective as DD that increases the connectivity between two subsets.

\subsection{Illustration on synthetic graphs}\label{sec:simulation}
In this section, we illustrate FA-GNN linking strategies on synthetic graphs and observe the GNN's response in terms of statistical parity difference ($SPD$).
We consider graphs of $4\,000$ nodes divided equally into four subsets of size $1\,000$, in the order $y_0s_0$, $y_0s_1$, $y_1s_0$, and $y_1s_1$.
Note that this uniform group distribution is for illustration purposes. We later illustrate the strategies on empirical datasets, where the groups are unbalanced.

\para{Generating node features.}
For each node, we generate $10$ features based on its class membership (for illustration purposes, node features are independent of group memberships in this setup). 
This results in a feature matrix $X \in \real{4\,000 \times 10}$, where $X(u,k)$ is the $k$-th feature of node $u$.
For a node $u \in y_is_j$, we sample 10 features from a normal distribution $X(u,k) \sim \mathcal{N}(\mu_{y_i}(k),\sigma^2)$ where $1 \leq k \leq 10$.
$\mu_{y_i} \in [-1,+1]^{10}$ is the mean vector sampled independently at random from $[-1,+1]$.
In our settings, we set $\sigma^2 = 0.5$.


\para{Generating the clean graph.}
To build our graphs, we use the stochastic block model~\cite{holland1983stochastic} (SBM) on the four subsets.
This requires a symmetric density matrix $M \in [0,1]^{4 \times 4}$ as an input.
As classical GNNs are typically deployed on homophilic graphs~\cite{zhu2020beyond,hussain2021interplay}, we follow a setup where the clean graph is homophilic w.r.t. labels.
To that end, we set the intra-label edge density to $0.004$ and the inter-label edge density to $0.0016$.
This preserves an appropriate level of sparsity and homophily in the graph.
Lower intra-label densities would increase the variance in neighborhood sampling. 
Besides, lower inter-label densities could result in extreme homophily, which makes the classification task trivial for the GNN model~\cite{zhu2020beyond,ma2021homophily}.


\para{Generating the perturbed graph.}
We apply the four strategies DD, DE, ED, EE on subset $A = y_1s_1$.
Let subset $B \in \mathcal{S}$ be the other subset to which we connect nodes of $y_1s_1$.
Let $|E|$ be the number of edges in the clean graph, which we sample using SBM as explained above.
For a perturbation rate $\delta$, we add $\delta |E|$ edges between random nodes from $A$ and random nodes from~$B$.
Note that in these illustrative experiments, the labels of all nodes are available to the attacker.
Hence, the attacker does not utilize a surrogate GCN.


\para{Evaluation.} 
After building the graph, we train a graph convolutional network~\cite{kipf2016semi} (GCN) model for node classification.
For the training process, we randomly sample $|V_l| = 20$ nodes to be in the training set, while the rest of the nodes are in the test set.
We repeat each trial for $100$ independent runs, implying a different graph structure, node features and GCN weight initialization in each run.

\para{Results.}
In Fig.~\ref{fig:linking} (bottom), for each strategy, we report the statistical parity difference after training the GCN across varying perturbation rates $\delta \in [0.05,0.3]$ with steps of $0.025$.
We find that 
\begin{enumerate*}[label=(\arabic*)]
  \item DD strategy increases $SPD$,
  \item DE strategy fluctuates around $0$,
  \item ED strategy also fluctuates around $0$ without a significant influence on $SPD$, and
  \item EE strategy decreases $SPD$ but at a lower rate than DD.
\end{enumerate*}
In addition, the DD strategy reaches a peak and then drops with more perturbations.
To address this and to reflect on our qualitative analysis in~\ref{sec:theoretical-analysis}, we report the changes in error rates from Equation~\ref{eq:sp-confusion} in our experiment.

\begin{figure}
    \centering
    \includegraphics[width=\linewidth]{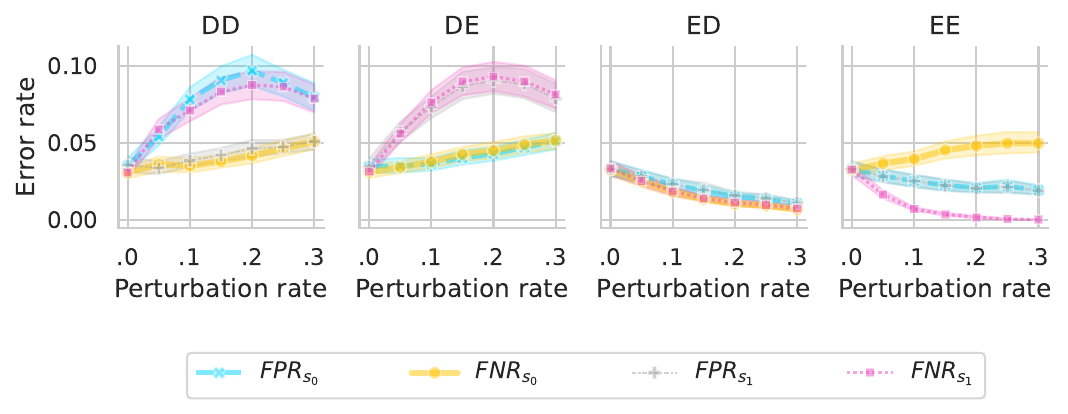}
    \caption{Error rates on synthetic graphs.
    Each figure corresponds to a linking strategy applied on subset $y_1s_1$ in a synthetic homophilic graph.
    $FPR_{s_j}$ and $FNR_{s_j}$ are the false positive and negative rates on subset $s_j$.
    Inter-label link injection (DD and DE) increase error on the involved subsets ($FNR_{s_1}, FPR_{s_0}$ with DD and $FNR_{s_1},FPR_{s_1}$ with DE).
    Conversely, intra-label link injection (ED and EE) decrease the error rates on the involved subsets ($FNR_{s_1},FNR_{s_0}$ with ED and $FNR_{s_1}$ with EE).
    In conclusion, involved subsets show a significant change in error rates with the attack. Meanwhile, the change in error rates on uninvolved subsets is particularly negligible for DD and DE.
    }
    \label{fig:synth-err}
\end{figure}

\para{Monitoring error rates.}
\label{sec:error}
Here, we show how error terms change with an increasing attack perturbation rate.
In Fig.~\ref{fig:synth-err}, we show the false positive/negative rates for each strategy with an increasing perturbation rate (averaged over 100 random seeds).
For group $s=j$, the false positive rate is $FPR_{s_j}=FP_{s_j}/n_{s_j}$ while the false negative rate is $FNR_{s_j}=FN_{s_j}/n_{s_j}$.
On average, involved terms increase for DD ($FPR_{s_0}, FNR_{s_1}$) and DE ($FPR_{s_1}, FNR_{s_1}$), and decrease for ED ($FNR_{s_0}, FNR_{s_1}$) and EE ($FNR_{s_1}$).
These changes are in line with our assumptions in Section~\ref{sec:theoretical-analysis}.

Uninvolved terms increase at a lower rate for DD and DE.
This particularly explains why, even by neglecting the changes in these terms, our analysis could approximate the final result of $SPD$ (Fig.~\ref{fig:linking} - Bottom).
For ED, uninvolved terms ($FPR_{s_0}, FPR_{s_1}$) decrease as well, which is expected due to better separation between the two classes.
For EE, $FNR_{s_0}$ also increases, which also reduces $SPD$.
We attribute this to the lower influence of the links between $y_1s_0$ and $y_1s_1$ as the latter attain high node degrees~\cite{hussain2021structack}.
This makes nodes with $y=0$ have a relatively high influence on $y_1s_0$ nodes and hence more false negatives of $s_0$.

Finally, for $DD$ strategy, we notice a drop at higher perturbation rates for both $FPR_{s_0}$ and $FNR_{s_1}$, which explains the observed drop in $SPD$ in Fig.~\ref{fig:linking} - Bottom.
A possible interpretation for the drop in $FPR_{s_0}$ and $FNR_{s_1}$ is that the subsets $y_0s_0$ and $y_1s_1$ are approaching extreme heterophily, which makes these subsets have distinct neighborhood distributions.
This case is well-studied in~\cite{ma2021homophily} and labeled as ``good heterophily'' where GNNs achieve lower error rates (note that $\delta=0.3$ represents a remarkably high perturbation rate). 

In conclusion, monitoring the error rates supports the assumptions for our analysis and opens a space for more exploration that we leave for future research.
Note that in this paper, we present our analyses for homophilic graphs.
However, we also performed our analysis and experiments on heterophilic and random graphs, and we found that DD and EE remain effective in attacking GNN's fairness also on these types of graphs (Appendix~\ref{app:theory}). 


\section{Experiments}\label{sec:experiments}


In this section, we evaluate FA-GNN on three real-world social network datasets.
Specifically, we aim to explore:
\begin{enumerate}
    \item \textbf{Effectiveness:} How do the proposed FA-GNN strategies perform in degrading the fairness of the GNN models?
    \item \textbf{Efficiency:} What is the impact of the amount of labeled data available to the attacker on the fairness results? 
    \item \textbf{Comparative analysis:}
    How do existing heuristics and accuracy-targeting attack methods influence GNN's fairness compared to FA-GNN?
    Does FA-GNN (unintentionally) degrade accuracy in comparison?
\end{enumerate}

Next, we introduce the datasets and baselines, followed by the experimental settings and results.




\para{Datasets.}
We conduct experiments on the following three empirical datasets. 1) \emph{Pokec\_z} and 2) \emph{Pokec\_n} \cite{dai2021say} are two social networks sampled from Pokec \cite{takac2012data}, a popular social network in Slovakia. Users in Pokec\_z and Pokec\_n belong to two major regions of different provinces. User features include gender, age, hobbies, interests, etc.  
The classification task for both datasets is to predict the working field of users with the user's region as the sensitive attribute. 
Pokec datasets do not reveal the specific working field or region names, so we only refer to their binary values.
3) \emph{DBLP} \cite{tang2008community} is a computer science bibliography network that contains papers from 4 areas (Database, Data Mining, Machine Learning, and Information Retrieval). Each area contains 5 representative conferences. We extract all the papers and authors in these 20 conferences, and construct a coauthor network based on co-author relationships. We match the author's gender as the sensitive attribute based on~\cite{jadidi2018gender}, and we construct the authors' features based on the words from their published papers' titles, which reveal their research interests. The classification task is to predict the research area of the author (in \textit{Database} or not).
We list the detailed statistics of the datasets in Table~\ref{tab:empirical-dataset}. 


\begin{table}[b]
\centering
\small
\caption{Statistics of the empirical datasets.}
\label{tab:empirical-dataset}
    \begin{threeparttable}[b]
    \begin{tabular}{@{}lrrrlll@{}}
    \toprule
    Dataset                      & Pokec\_z & Pokec\_n & DBLP     \\ \midrule
    \# of nodes                  & 67,435   & 66,082   & 20,111   \\
    \# of edges                  & 617,765  & 516,784  & 57,508  \\
    Feature dimension    & 274      & 263      & 2,491      \\
    \% of nodes in $y_0s_0$   & 30.88\%  & 35.46\%  & 54.21\%  \\
    \% of nodes in $y_0s_1$  & 15.57\%  & 15.81\%  & 12.58\%  \\
    \% of nodes in $y_1s_0$  & 33.55\%  & 32.40\%  & 27.63\% \\
    \% of nodes in $y_1s_1$  & 20.00\%  & 16.33\%  & 5.58\%   \\ 
    Intra-label density* ($\times 10^{-4}$) & 4.7 & 4.8 & 12.5\\
    Intra-group density ($\times 10^{-4}$) & 7.0 & 6.9 & 8.3\\\bottomrule
    \end{tabular}
    \begin{tablenotes}
     \item[*] The intra density is the number of intra-edges divided by the number of all possible edges.
   \end{tablenotes}
  \end{threeparttable}
\end{table}



\para{Fairness attack.}
We compare four adversarial linking strategies of FA-GNN, i.e., DD, DE, ED, EE, as denoted in Section~\ref{sec:strategy}.
We apply these strategies on all four subsets in $\mathcal{S}$.
Please note that some strategy-subset combinations are equivalent.
For example, DD on $y_1s_1$ is equivalent to DD on $y_0s_0$, and they both mean adding edges between random nodes of $y_1s_1$ and $y_0s_0$.
In such cases, we do not report duplicates, which totals up to $10$ possible FA-GNN attacks.

\para{Baselines.}
To validate whether existing attacks on accuracy can (unintentionally) degrade fairness, and to compare the performance of FA-GNN, we consider the following attack methods:
\begin{itemize}
    \item \textbf{Random}: A baseline attack method that randomly adds edges to the network.
    \item \textbf{DICE} \cite{DBLP:journals/corr/WaniekMRW16}: An accuracy-targeting baseline attack method that connects nodes with different labels and disconnects nodes with the same labels. 
    \item \textbf{PR-BCD} \cite{geisler2021robustness}: A SotA accuracy-targeting attack method via  adding/removing edges based on Randomized Block Coordinate Descent (R-BCD), which scales to large graphs.
\end{itemize}
Note that the space requirements of attacks such as~\cite{zugner2019adversarial,wu2019adversarial,chen2020mga,xu2019topology} are prohibitive given the sizes of our datasets, so we only consider scalable attacks for our comparison.

\para{Metrics.} We analyze FA-GNN's performance based on the three fairness metrics as introduced in Section~\ref{preliminaries}, i.e., \emph{statistical parity difference} ($SPD$), \emph{equality of opportunity difference} ($EOD$), and \emph{equalized odds difference} ($EQD$).
The farther these metrics are from 0, the more unfair the results are.

\para{Victim GNN models.}
To comprehensively evaluate the effectiveness of the proposed attack strategies, we compare the following GNN models: 1) \emph{general GNN models}, i.e., \emph{GCN}~\cite{kipf2016semi}, \emph{GAT}~\cite{velivckovic2018graph}, and \emph{GraphSAGE}~\cite{hamilton2017inductive}, and 2) \emph{fairness-enhancing GNN models}, i.e., \emph{FairGNN}~\cite{dai2021say} and \emph{NIFTY}~\cite{agarwal2021towards}. 

\para{Surrogate GCN model}. In our setting, the attacker's knowledge of node labels is limited to the training set.
To obtain the labels of the remaining nodes, we use GCN as the surrogate model for label prediction, as introduced in Section~\ref{sec:fagnn}. 
This surrogate GCN is independent of the victim GNN model for node classification.

\para{Experimental settings.}
We use GNN models of two layers with 64-dimensional hidden layers.
For GAT, we use one layer with 2 attention heads on the Pokec datasets, and 2 layers with 8 attention heads on DBLP.
For GraphSAGE, we use the mean aggregator for node feature aggregation.
For FairGNN, we use GCN as the base model, and the number of sensitive attributes for the sensitive attribute estimator is set to 200.
The hyperparameters $\alpha$ and $\beta$ in FairGNN for balancing the loss functions are set to 2 and 0.1, respectively.
For the victim GNN model training, we randomly pick 50\%, 25\%, 25\% of labeled nodes as training set, validation set, and testing set.
For the attacker, the available labeled nodes are the same as the victim GNN's training set.
We increase the attack perturbation rate from 0.05 to 0.3 with a 0.05 increment.
We train all models on each of the perturbed networks for 500 epochs and choose the parameters with the best node classification accuracy on the validation set.
We report the mean and standard deviations of fairness metrics on the test set on 5 independent runs.

\subsection{Effectiveness of attacks} \label{sec:effe}

\begin{figure}
    \centering
    \includegraphics[width=0.49\textwidth]{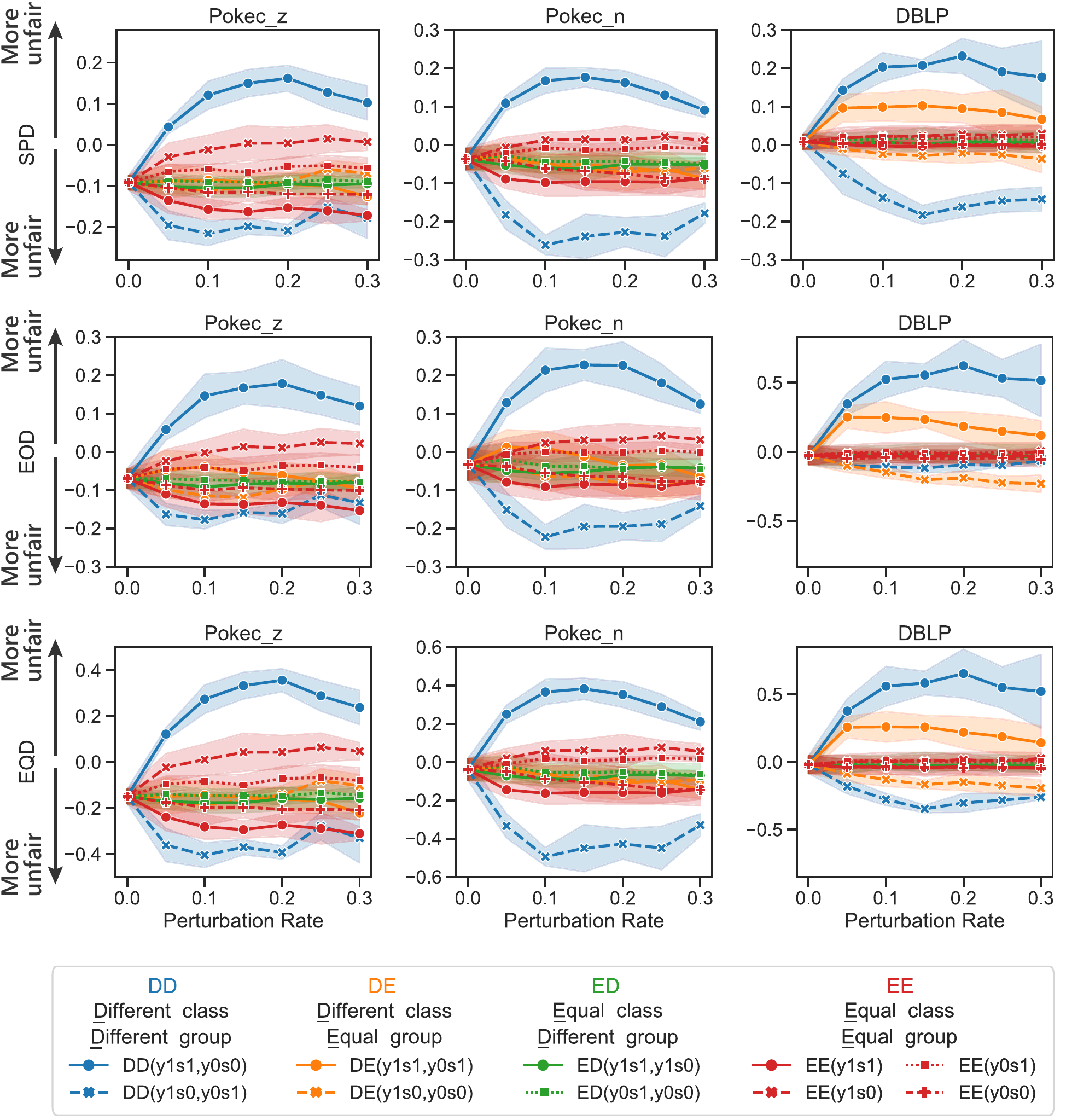}
    \caption{FA-GNN evaluation with GCN on empirical datasets.
    We show the three fairness measures in terms of increased perturbation rate, i.e., statistical parity difference (SPD), equality of opportunity difference (EOD), and equalized odds difference (EQD). The arrows show where the result becomes more unfair.
    Each line refers to a linking strategy between two subsets (or within one subset for EE).
    We notice that DD achieves the most significant shift in all fairness metrics.
    EE achieves a slight change on Pokec datasets, while DE achieves a slight change on DBLP.
    The figure shows that DD is the most effective linking strategy, and that the results of EE and DE differ based on the input graph.
    }
    \label{fig:all-fairness-gcn}
\end{figure}

\para{Fairness metrics.}
First, we evaluate the attack performance on the GCN model based on three signed fairness metrics.
We present the results in Fig.~\ref{fig:all-fairness-gcn}, where a perturbation rate of 0 represents the clean graph.
With the perturbation rate increasing, the deviation of the curves away from the center line of zero (absolute fairness) indicates increased unfairness.
Based on the results, we make the following observations:

    On Pokec\_z/n we see that out of the four strategies, DD and EE result in a consistent change in $SPD$, while DE and ED have no significant influence.
    These observations are also valid for DBLP, except for EE and DE. 
    In particular, EE appears to have a minor effect on DBLP, while DE causes a clear shift in $SPD$. We attribute this to different generative processes of the original networks. For example, Pokec\_z/\_n have a relatively high intra-group edge density, while DBLP has a high intra-label edge density (Table~\ref{tab:empirical-dataset}). Hence, adding edges among the same labels may show different performance in the empirical networks. 
    Overall, \emph{DD is the most effective on all three datasets}: DD($y_1s_1$,$y_0s_0$) increases $SPD$ while DD($y_1s_0$,$y_0s_1$) decreases $SPD$.
    Similar to the synthetic graphs, the performance of DD strategy drops for high perturbation rates.
    We also attribute that to the high heterophily rate (or ``good heterophily'') attained at this stage.
    For example, DD($y_1s_1,y_0s_0$) decreases the label homophily ratio of nodes belonging to $y_1s_1$ in DBLP from $0.43$ at perturbation rate $0.10$ to $0.34$ at perturbation rate $0.30$.
    As~\cite{ma2021homophily} suggests, we assume that at this level, $y_1s_1$ nodes get a more distinct neighborhood and hence a lower error rate.
    
    Our analysis (Sections~\ref{sec:theoretical-analysis} and~\ref{sec:simulation}) was limited to statistical parity difference.
    Fig.~\ref{fig:all-fairness-gcn} shows that FA-GNN also degrades fairness in terms of $EOD$ and $EQD$.
    This shows the effectiveness of FA-GNN method in degrading GNN's fairness on various fairness metrics.
    

    
    

\begin{figure}
    \centering
    \includegraphics[width=0.49\textwidth]{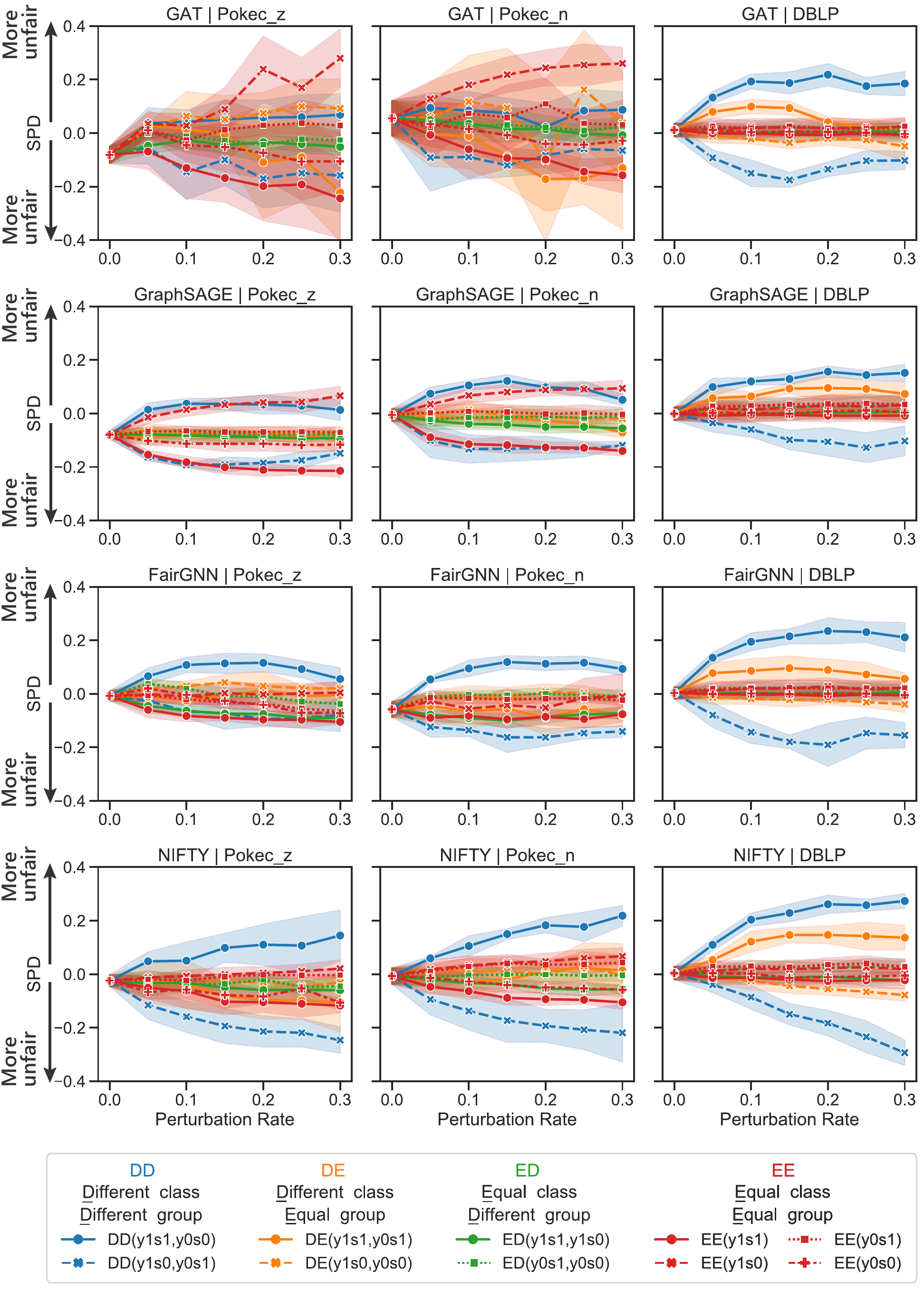}
    \caption{FA-GNN evaluation with different GNN models on empirical datasets.
    We show the statistical parity difference ($SPD$) in terms of increased perturbation rate.
    The arrows show where the result becomes more unfair.
    Each line refers to a linking strategy between two subsets (or withing one subset for EE).
    DD achieves the most significant shift in $SPD$. 
    }
    \label{fig:dblp-parity-all-gnns}
\end{figure}


\para{Various victim GNN models.}
Next, we compare the attack performance of statistical parity difference on four more victim GNN models, i.e., GAT, GraphSAGE, FairGNN, and NIFTY.
We present the results on the three datasets with varying perturbation rates in Fig.~\ref{fig:dblp-parity-all-gnns}. 
The results show that FA-GNN is effective in attacking the fairness of various GNN models, where \emph{DD strategy consistently increases unfairness}. 
DD strategy can even degrade the fairness of the fairness-enhancing FairGNN and NIFTY models. We also observe that the attack performance on GAT show a large standard deviation on the Pokec\_n/z. This could be due to the instability brought by the small number of attention heads and the complex hyperparameters tuning required for training the model.

Summing up, the results show that \emph{FA-GNN is effective in terms of different fairness metrics on different GNN models}.
Hence, FA-GNN degrades the fairness of various GNN models 
without prior knowledge on the victim GNN model.
It is also worth noting that for the FairGNN and NIFTY models, which are intentionally designed to enhance fairness, FA-GNN is still effective in degrading the fairness. Especially for NIFTY with a robust design against random perturbations on node sensitive attributes and network structure, FA-GNN still shows a significant increase in SPD for all datasets.  
This observation supports not only \emph{effectiveness} but also \emph{robustness} of the attack in degrading GNNs' fairness even under certain fairness-enhancing GNN models.

\para{Attack outcome.}
\label{sec:attack-outcome}
In Figures~\ref{fig:all-fairness-gcn} and~\ref{fig:dblp-parity-all-gnns} we observe that GNNs generate unfair prediction results even for the clean graphs.
Note that for all the fairness metrics, any deviation from 0 irrespective of the sign, indicates unfairness with respect to one group or the other. 
Interestingly, the fairness results on the clean graph is already an indicator of how the attack unfolds with more perturbations.
This means that it is easier for the attack to exacerbate the already existing disadvantage rather than attacking the advantaged group. 
For example, GCN in Fig.~\ref{fig:all-fairness-gcn} shows a negative $SPD$ on clean Pokec\_z, so it's easier for strategies that reduce (signed) $SPD$ to make fairness even worse.
In particular, DD($y_1s_0$,$y_0s_1$) on Pokec\_z reaches $|SPD|\sim0.2$ at a low perturbation rate of $0.1$ compared to DD($y_1s_1$,$y_0s_0$) that reaches a similar $|SPD|$ at a higher perturbation rate of $0.2$.

\subsection{Efficiency of attacks}
\begin{figure}
    \centering
    \includegraphics[width=0.5\textwidth]{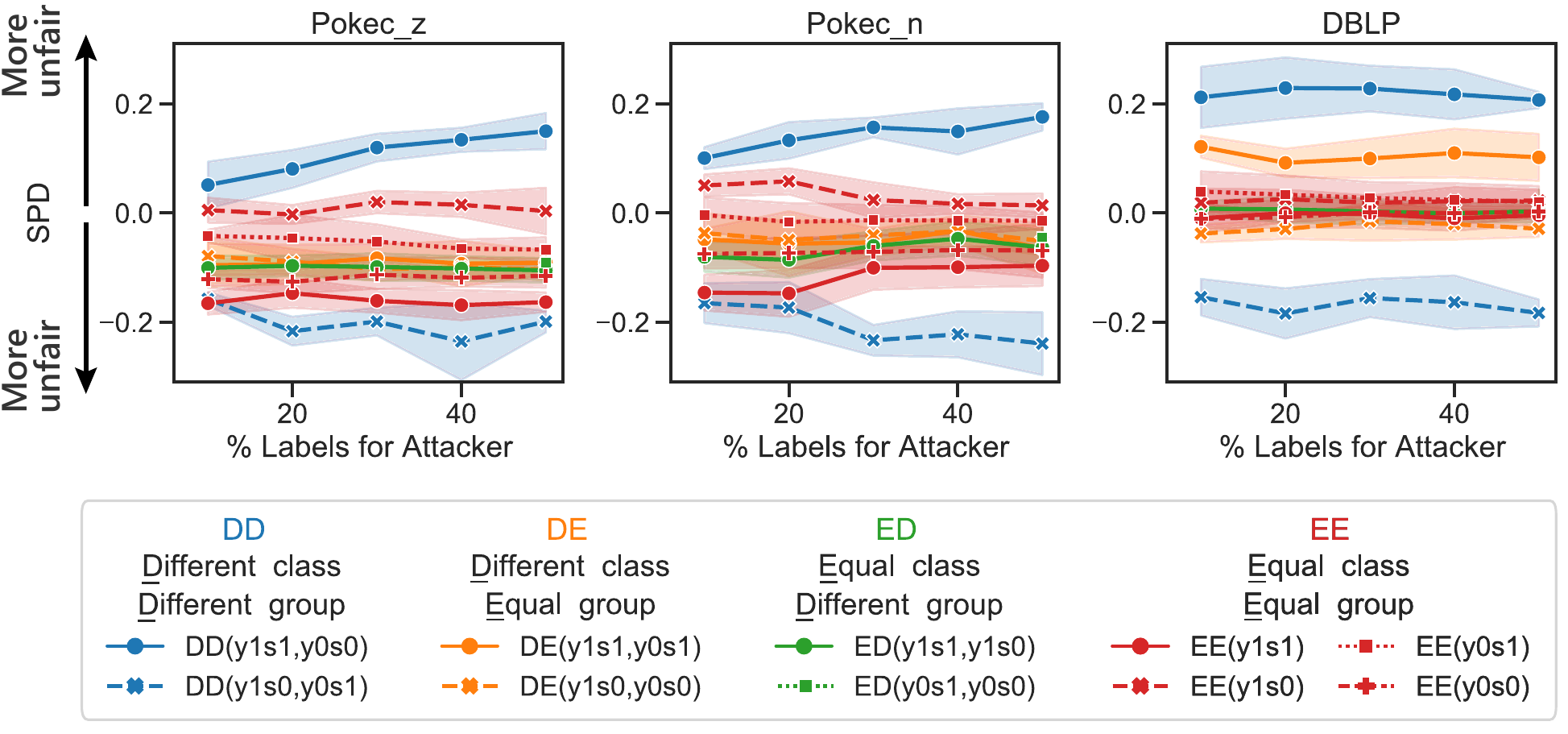}
    \caption{FA-GNN efficiency analysis.
    The x-axis shows the percentage of the labeled nodes available to FA-GNN.
    The y-axis shows the statistical parity difference (SPD) of a GCN model trained on the perturbed graph. 
    It shows that FA-GNN degrades fairness even at a low percentage of labels.
    More labeled nodes increase the influence of FA-GNN on SPD, especially with DD strategy.
    }
    \label{fig:gcn-parity-train-set-increase}
\end{figure}
We evaluate the efficiency of FA-GNN in terms of the amount of labeled data available to the attacker. Note that in our setting, the attacker has access to ground truth labels only for a subset of nodes. 
To that end, we consider varying percentages of labeled nodes available to the \emph{surrogate GCN}: from 10\% to 50\% with a 10\% increment.
Recall that FA-GNN uses these labeled nodes to train a surrogate model and predict the labels of the unlabeled nodes.
We keep the ratios of labeled data for the \emph{victim GNN model} unchanged: 50\% training, 25\% validation and 25\% testing.
The perturbation rate is fixed at $0.15$ and GCN is considered as the victim model which is trained on the perturbed network. 
We report the statistical parity of the GCN's predictions in Fig.~\ref{fig:gcn-parity-train-set-increase}.

From the results, we observe that,
1) \emph{efficiency:}
with a small percent of available labeled data, i.e., 10\%, FA-GNN already degrades the GCN's fairness to a large extent.
This indicates that FA-GNN can efficiently harm GNN's fairness with very limited access to the information on true labels;
2) \emph{influence of labeled data:}
there is no obvious increase in $SPD$ with increasing number of labeled nodes, except for DD strategy on Pokec\_z/n, where the performance slightly improves with more labeled nodes.
We also observe similar results for $EOD$ and $EQD$. This demonstrates that the availability of more labeled nodes does not necessarily lead to more successful fairness attacks. 



\subsection{Comparative analysis of attacks}\label{sec:comparative}
\begin{figure}
    \centering
    \includegraphics[width=0.49\textwidth]{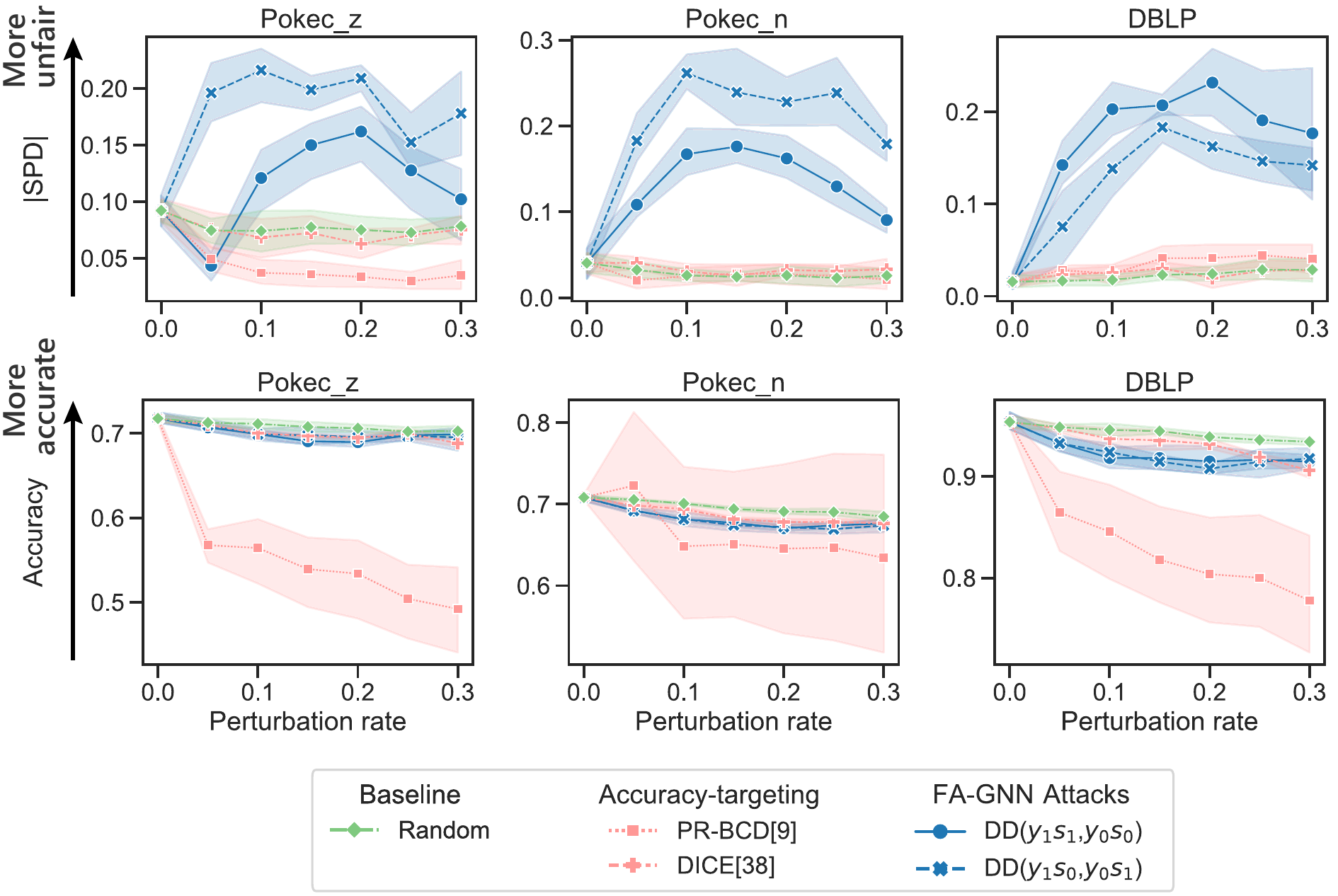} 
    \caption{Comparative evaluation of FA-GNN against baselines. We compare FA-GNN to existing adversarial attacks on empirical datasets in terms of absolute statistical parity ($|SPD|$) and classification accuracy with increased perturbation rate. The result shows that existing attacks do not have a significant impact on fairness compared to FA-GNN. In terms of accuracy, FA-GNN does not have a significant effect compared to the SotA PR-BCD~\cite{geisler2021robustness}.}
    \label{fig:all-parity-gcn-baselines} 
\end{figure}

We evaluate a baseline attack method (Random) and two accuracy-targeting attack methods (DICE, PR-BCD) in degrading GCN's fairness.
We compare the results with FA-GNN's DD strategy, i.e., $y_1s_1$-DD and $ y_1s_0$-DD.
We report both absolute statistical parity difference $|SPD|$ and node classification accuracy on empirical datasets in Fig.~\ref{fig:all-parity-gcn-baselines}. 
Higher $|SPD|$ implies more unfair predictions.

From the results, we observe that:
1) \emph{fairness:} Random and accuracy-targeting attack methods have only little impact on the fairness results in terms of $|SPD|$ score. Compared to the baselines, FA-GNN strategies are much more effective in  degrading the fairness results of GCN.
We also make similar observations for $|EOD|$ and $|EQD|$;
2) \emph{model performance in accuracy:} While Random, DICE, and FA-GNN show no significant drop in accuracy, the SotA accuracy-targeting PR-BCD significantly degrades model performance with a large drop in accuracy.


Summing up, successful attacks on GNNs' accuracy do not automatically translate to attacks on GNNs' fairness.
Moreover, FA-GNN only has a marginal effect on the accuracy across all the datasets.
This indicates that these fairness attacks can maintain a relatively good performance of the GNN model, making them less likely to be detected by a defender (\emph{more deceptive}), at least based on model performance.

\section{Discussion}

In this section, we point out the limitations and discuss the implications of our results. 

\subsection{Limitations}
\para{Analyses.}
Our analysis was limited to statistical parity and hinges on assumptions that hold in some real-world networks.
In particular, we have assumed increasing error rates of involved subsets, and we have neglected uninvolved subsets.
We realize that this is an approximate analysis that is limited to some real-world scenarios.
We believe that a more generalized analysis can be reached in the future.
Further, we only analyze the attack's influence on the GNN fairness from the perspective of edge distributions.
However, several other factors can impact fairness, such as label and sensitive attribute distributions, and node features.
Nevertheless, the results of our experiments on synthetic and real-world data open up the opportunity of several questions about the interplay between GNN fairness and the graph structure.



\para{Alternate effective attacks.}
We base the attack strategies on structural perturbations through injecting edges.
Rewiring/deleting edges, injecting nodes and poisoning node features might also lead to effective attack strategies on fairness.
Besides, we focus on injecting adversarial edges based on sensitive attributes and labels.
These injection strategies could be extended by considering node feature similarity as a third contributing factor.
We believe that designing an attack that hurts fairness maximally requires separate research efforts, as our paper focuses on introducing the problem of fairness attacks on GNNs. 

\subsection{Implications}


\para{Deceptiveness of attacks.}
Our results show that fairness attacks can significantly degrade prediction fairness without having a significant impact on prediction accuracy.
This demonstrates that checking for model accuracy might not be sufficient for detecting fairness attacks.
If model fairness is not monitored, such attacks can go unnoticed, exacerbating the consequences for certain groups.

\para{The trap of inter-group links.}
Our analysis shows the vulnerability of GNNs both to fairness attacks and to non-adversarial edge creation in the network.
The underlying generative process of edges might conform to a certain linking strategy (such as DD), which could also result in unfair predictions.
For example, it is plausible that encouraging links among, e.g. different demographic groups, can lead to an improvement of some notion of fairness.
In fact, some related work assumes or hints that inter-group linking promotes fairness of predictions~\cite{spinelli2021fairdrop,masrour2020bursting}.
Counter-intuitively, our results show that such linking might increase the disadvantage of an already disadvantaged group.
This observation suggests that monitoring edge creation processes in networks might be warranted when aiming to maintain fairness of predictions. It also highlights the challenges of conflicting notions of fairness, i.e. where increasing some notion of fairness may degrade fairness of predictions as a side effect.

\para{Defense against FA-GNN.}
Besides analyzing fairness attacks on GNNs, our work highlights the urgency of developing counter measures against such attacks.
As a first step, we have evaluated state-of-the-art fairness-enhancing frameworks (FairGNN and NIFTY) and showed that they cannot defend against FA-GNN.
We further evaluate DropEdge~\cite{rong2019dropedge} and a modified version of it that drops random DD edges.
We do not see any significant improvement in fairness compared to GCN on the empirical datasets. 
Therefore, we believe that defending against FA-GNN requires better-tailored designs.

\section{Related work}

In this work, we build on the following streams of research.

\para{Adversarial attacks on graphs.} 
Adversarial attacks on graphs~\cite{jin2020adversarial} (and GNNs in particular~\cite{zugner2018adversarial,zugner2019adversarial,xu2019topology,hussain2021structack,ma2020towards}) have been mainly concerned with reducing the model accuracy.
The key idea involves designing structural perturbations (addition or removal of edges or nodes) on the underlying network or manipulating the node features in order to corrupt the accuracy of predictions.
The models mostly differ with respect to the type of attack, type of perturbation, attacker's goal and attacker's knowledge~\cite{jin2020adversarial}. 
However, adversarial attacks targeting fairness of prediction results, to the best of our knowledge, have largely remained unexplored.  

\para{Fairness in graph learning.}
With machine learning algorithms being deployed across different real-world applications involving humans, bias in algorithmic decision-making has received increasing attention. 
As graph models have gained popularity in research and industry, a body of work has developed to reduce the bias in node representation and predictions on several graph mining tasks.
In this direction, FairWalk~\cite{rahman2019fairwalk} introduces a modified random walk embedding approach to reduce the bias in link prediction.
\cite{bose2019compositional} propose an adversarial training based method to counter bias in graph embeddings, albeit without node features. 
\cite{ma2022learning} also propose to learn fair node embeddings with graph counterfactual fairness.  
For GNNs in particular, FairGNN \cite{dai2021say} proposes a general adversarial learning based setup to mitigate bias, given limited access to the sensitive attribute information.
NIFTY \cite{agarwal2021towards} considers the notion of counterfactual fairness and introduces a framework to learn both fair and stable node representation, which aims to ensure the stability of inference results to perturbations of sensitive attribute values.
Finally, \cite{ma2021subgroup} provides insights on the relation between the network structure and the subgroups marginalized by GNNs.

\para{Attacks on fair machine learning.}
A growing body of work 
has been studying the robustness of more traditional machine learning models to attacks on fairness~\cite{mehrabi2020exacerbating,nanda2021fairness,solans2020poisoning,chhabra2021fairness}.
Solans et al.~\cite{solans2020poisoning} introduced the idea of designing adversarial attack strategies targeting fairness of machine learning models. The authors develop a gradient-based poisoning attack aimed at introducing classification disparities among different groups in the data.
Mehrabi et al.~\cite{mehrabi2020exacerbating} propose two families of data poisoning attacks targeting fairness (anchoring and influence attacks), which inject poisoned data points aiming to degrade fairness. 
Chhabra et al.~\cite{chhabra2021fairness} present an adversarial attack strategy targeting the fairness of clustering algorithms. 

To the best of our knowledge, our work is the first to investigate adversarial attacks targeting the fairness of GNNs.
Arguably \cite{mehrabi2020exacerbating}, and \cite{solans2020poisoning} are the most related to our work, as they aim to design data poisoning attacks targeting fairness but not on graph data. 

\section{Conclusion}

In this paper, we presented adversarial attack strategies that target the fairness results of graph neural network (GNN) based node classifier models. 
We started by defining fairness attacks on GNNs and describing adversarial strategies that degrade fairness.
Then, we provided a qualitative analysis to understand the consequences of these adversarial strategies on the fairness of GNN predictions.
We illustrated these consequences on synthetically generated graph datasets.
Our evaluations on empirical datasets showed that fairness attacks significantly degrade the fairness of node classification results without a significant drop in accuracy.
We further empirically demonstrated that the presented strategies could be successfully extended to several GNN models such as GCN, GAT, GraphSAGE, and even inherently fair models like FairGNN and NIFTY.
Although developed with statistical parity in mind, the proposed strategies can have similar effects on other fairness metrics like equality of opportunity and equalized odds.
Our work demonstrates the vulnerability of GNNs to simple structural perturbation based adversarial attacks on fairness.
Designing methods that improve the fairness of GNNs while being robust against such adversarial attacks should be a pressing concern for future research.

\bibliographystyle{IEEEtranS}
\bibliography{sample-base}
\clearpage
\appendices



\section{Model implementations}
\label{sec:reproducibility}
\para{GNN models.}
For GCN, GAT, and GraphSAGE, we adopt the implementations provided in the DGL package~\cite{wang2019dgl}. For FairGNN~\cite{dai2021say} and NIFTY~\cite{agarwal2021towards}, we adopt the implementation provided by the authors.
For all baselines, we use the same random seeds to produce the same training, validation, and test sets. 

\para{Graph Attack Baselines.}
For Random and DICE, we adopt the implementation provided in the deeprobust package~\cite{li2020deeprobust}. For PR-BCD \cite{geisler2021robustness}, we use the implementation provided by the authors.

\para{Hyperparameters.}
The hyperparameters $\alpha$ and $\beta$ in FairGNN for balancing the loss functions are set to 2 and 0.1, respectively. For the victim GNN model training, we randomly pick 50\%, 25\%, 25\% of labeled nodes as training set, validation set, and testing set. We train all GNN models both on the clean networks and attacked networks with a learning rate of 0.001. For GAT on Pokec datasets, we use a dropout of 0.8 and a weight decay of 1.9e-2, for all the other models, we set dropout as 0.6, and weight decay as 5e-4. For the surrogate GCN, we use two hidden layers with 16 dimensions for the first hidden layer, and we train the model for 500 epochs with a learning rate 0.01, weight decay 5e-4, and droupout 0.5.

\section{Extended analysis}
\label{app:theory}

\begin{figure}[t]
     \begin{center}
    \includegraphics[width=\linewidth]{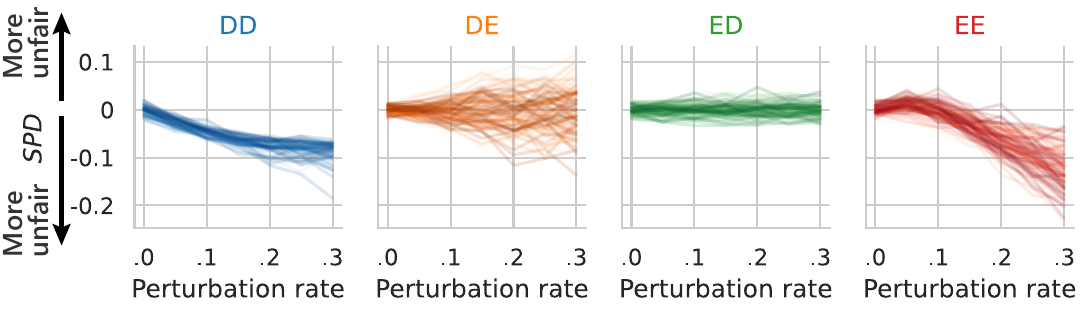}
    \includegraphics[width=\linewidth]{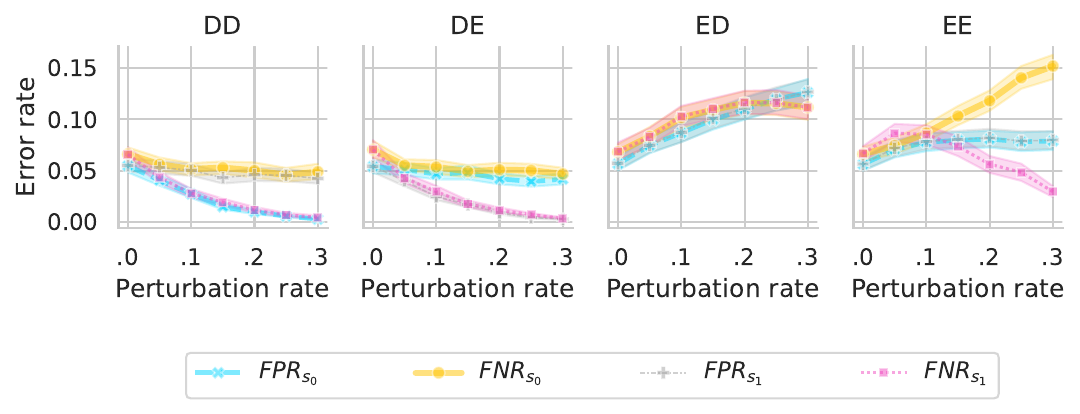}
    \caption{Fairness attacks on heterophilic synthetic graph.
    We show the statistical parity difference ($SPD$) (top) and error rates (bottom) of FA-GNN strategies on heterophilic synthetic graph.
    (Top) Each figure corresponds to a linking strategy applied on subset $y_1s_1$.
    While DE and ED do not have a targeted influence on $SPD$, DD decreases $SPD$, and EE only increases it at the beginning.
    (Bottom) $FPR_{s_j}$ and $FNR_{s_j}$ are the false positive and negative rates on subset $s_j$.
    Attacks that decrease label homophily (DD and DE) \textit{decrease} error on the involved subsets ($FNR_{s_1}, FPR_{s_0}$ with DD and $FNR_{s_1},FPR_{s_1}$ with DE).
    Attacks that increase label homophily (ED and EE) \textit{increase} the error rates on the involved subsets ($FNR_{s_1},FNR_{s_0}$ with ED and $FNR_{s_1}$ with EE at the beginning).
    This shows that for heterophilic graphs, DD and EE are still effective attacks.
    }
    \label{fig:synth-results-het}
         
     \end{center}
 \end{figure}

\para{Results on heterophilic graphs.}
In our analysis (Section~\ref{sec:theoretical-analysis}), we assume that the linking of two subsets from different classes increases the prediction error rates on these subsets.
Here, we aim to relax this assumption by considering a case where linking different classes decreases the prediction error rates on both subsets.
This scenario is possible if the original network does not have structural features that help the classifier distinguish between two subsets.
A specific example is when two subsets of different classes have a similar neighborhood distribution in the clean graph~\cite{ma2021homophily}.
Then such a linking strategy makes two subsets of nodes more distinguishable for the classifier, hence improving the prediction on both subsets.
For linking subsets of the same class, we consider that the error rate will increase as they will become less distinguishable.
A hypothetical case where this can happen is when the same-class subsets become more similar to a subset of another class in terms of neighborhood distribution.
Having similar neighborhood distributions can make the respective subsets indistinguishable~\cite{ma2021homophily}.
We reconsider our strategies and analyze the statistical parity difference.
$$SPD = \frac{FP_{s_0}-FN_{s_0}}{n_{s_0}} - \frac{FP_{s_1}-FN_{s_1}}{n_{s_1}} + \frac{n_{y_1s_0}}{n_{s_0}}-\frac{n_{y_1s_1}}{n_{s_1}}$$
Again, we apply the following strategies to $y_1s_1$
\begin{enumerate}
    \item[(DD)] $\Delta FP_{s_0} < 0, \Delta FN_{s_1} < 0 \Rightarrow \Delta SPD < 0$ and the attack is in favor of $s_1$.
    \item[(DE)] $\Delta FP_{s_1} < 0, \Delta FN_{s_1} < 0$ the sign of $\Delta SPD$ still depends on the error rate of each subset.
    \item[(ED)] $\Delta FN_{s_0} > 0, \Delta FN_{s_1} > 0$ the sign of $\Delta SPD$ still depends on the error rate of each subset.
    \item[(EE)] $\Delta FN_{s_1} > 0 \Rightarrow \Delta SPD > 0$ and the attack is in favor of $s_0$.
\end{enumerate}
To simulate this scenario, we perform the same experiments on synthetic datasets described in Section~\ref{sec:simulation}.
However, in this setup, we start with a heterophilic graph: the edge density is $0.0016$ for subsets of the same class and $0.004$ for subsets of opposite classes.
Figure~\ref{fig:synth-results-het} shows the statistical parity difference and the error rates of this simulation and supports our hypotheses.
In this setup, EE only shows an increase in $FNR_{s_1}$ at the beginning.
This could be due to $y_1s_1$ obtaining a similar neighborhood distribution to one subset and then gaining its unique neighborhood distribution.

\para{Results on random graphs.}
On random graphs, we generally see similar results, except that error rates do not increase with same-class linking (ED/EE). 

\end{document}